\documentclass[a4paper,twoside]{article}

\usepackage{epsfig}
\usepackage{subcaption}
\usepackage{calc}
\usepackage{amssymb}
\usepackage{amstext}
\usepackage{amsmath}
\usepackage{amsthm}
\usepackage{multicol}
\usepackage{pslatex}
\usepackage{natbib}
\usepackage{apalike}
\usepackage{algorithm2e}
\usepackage[bottom]{footmisc}
\usepackage{svg}
\usepackage{SCITEPRESS}     

\usepackage{xcolor}

\begin{document}

\title{Hierarchically Gated Experts for Efficient Online Continual Learning}

\author{\authorname{Kevin Luong\sup{1}\orcidAuthor{0009-0007-3362-9082} and Michael Thielscher\sup{1}\orcidAuthor{0000-0003-0885-2702}}
\affiliation{\sup{1}The University of New South Wales, Sydney, Australia}
\email{\{kevin.luong, mit\}@unsw.edu.au}
}

\keywords{Neural Networks, Machine Learning, Continual Learning, Mixture-of-Experts}

\abstract{Continual Learning models aim to learn a set of tasks under the constraint that the tasks arrive sequentially with no way to access data from previous tasks. The Online Continual Learning framework poses a further challenge where the tasks are unknown and instead the data arrives as a single stream. Building on existing work, we propose a method for identifying these underlying tasks\/: the \emph{Gated Experts (GE) algorithm\/}, where a dynamically growing set of experts allows for new knowledge to be acquired without catastrophic forgetting. Furthermore, we extend GE to \emph{Hierarchically Gated Experts (HGE)\/}, a method which is able to efficiently select the best expert for each data sample by organising the experts into a hierarchical structure. On standard Continual Learning benchmarks, GE and HGE are able to achieve results comparable with current methods, with HGE doing so more efficiently.}

\onecolumn \maketitle \normalsize \setcounter{footnote}{0} \vfill

\section{\uppercase{Introduction}}
\label{sec:introduction}
Continual Learning, also known as Lifelong Learning, addresses the challenge of creating AI able to continually learn and apply knowledge over long time spans \citep{Lifelong}. The biggest barrier to achieving such an AI is catastrophic forgetting, where previously learnt knowledge is lost due to interference when acquiring new knowledge. In practice, this is observed as a reduction in performance on existing tasks when learning a new task. 

In Continual Learning an agent is restricted from seeing past data samples yet must correctly model both past and present data. Whilst past data samples are always forbidden, in many continual learning settings the data is assumed to come from a series of separate distributions (tasks) with each sample labelled according to its distribution \citep{Kirkpatrick2017, GEM}. This task identity data is not available in the Online Continual Learning setting, which is currently gaining attention as a more useful extension to Continual Learning \citep{HCL, hihn2023hierarchically}.

Many Continual Learning techniques in both online and offline settings aim to mitigate interference between tasks, and thus catastrophic forgetting, by employing a set of experts such that each expert is associated with a single task \citep{ExpertGate, zhu2022tame}. The challenge, in online settings, is then in selecting the correct expert during both training and testing. We build on existing work in this area to propose our own method, which we name Gated Experts (GE). More specifically, our algorithm can be seen as an extension of Expert Gate \citep{ExpertGate} to the online setting by detecting task switches as statistically significant deviations in the training loss. We credit \citet{zhu2022tame} for the idea to use the training loss as a task switch signal. However, we track the loss and detect task switches in a different manner to address shortcomings in the existing approach, discussed in section \ref{subsec:TAME}.

The GE algorithm belongs to the class of expansion-based methods, where the number of parameters dynamically grows as needed to learn new tasks. In this area, the efficiency of an algorithm is often optimised by reducing the number of parameters, by pruning \citep{zhu2022tame} or other methods \citep{DEN, RCL}. We instead propose to organise the experts in a hierarchical manner, so that only a subset of experts are required for any given data sample. Thus, the inference time can be improved. We name this extension Hierarchically Gated Experts (HGE).

We evaluate GE on standard Continual Learning benchmarks and show that it is competitive with the state-of-the-art in Online Continual Learning. Then, we create new Continual Learning scenarios involving a mixture of datasets to which we apply both GE and HGE. We show that HGE is able to organise experts hierarchically, thus increasing efficiency, with little loss in accuracy.

In summary, our contributions are:
\begin{itemize}
    \item We present a novel method for task switch detection to address shortcomings in an existing approach.
    \item We apply this method on existing work to propose the GE algorithm and empirically show it to be competitive with the state-of-the-art in Online Continual Learning.
    \item We propose hierarchical organisation, unexplored in the literature thus far, as a means of improving the efficiency of expansion-based Continual Learning methods. Our HGE extension improves upon the efficiency of GE. We demonstrate the extension on bigger Continual Learning scenarios as a proof-of-concept.
\end{itemize}

The rest of the paper is structured as follows. We provide an overview of Continual Learning and current methods in Section \ref{sec:background}. Then, we present our approach to task switch detection and the GE algorithm in section \ref{sec:OCL}, and describe the HGE extension in section \ref{sec:HGE}. Our experimental setup and results are given in Section \ref{sec:experiments}, and we conclude with a summary and discussion of future work in Section \ref{sec:conclusion}.

\section{\uppercase{Background}}
\label{sec:background}
\subsection{Continual Learning}
There has been a wide array of approaches proposed for continual learning. We provide a broad overview of the continual learning landscape, but refer to surveys for a taxonomy as well as a more comprehensive coverage of these approaches \citep{Lifelong, wang2024survey}. There exists three major categories of continual learning methods: architectural, regularisation, and replay-based.

\textbf{Architectural approaches} aim to assign specific model parameters to specific tasks, thus mitigating catastrophic forgetting as each parameter is only fitted to one task. Often, there is a separate model, referred to as an expert, that is assigned to each task to be learnt \citep{zhu2022tame, CNDPM}. The Expert gate \citep{ExpertGate} algorithm creates a new expert for each task during training, as well as an \emph{autoencoder\/} to learn a representation for the task. During test time, each sample is forwarded to the expert corresponding to the autoencoder that best recreates the sample. Alternatively, there exist approaches where
the network is fixed but binary masks are learned to restrict each task to a subset of parameters \citep{WSN, supermasks}. In some cases, the task-specific parameters may not be stored and instead generated by a meta-model known as a hypernetwork (HN) \citep{HN}.

\textbf{Regularisation techniques} constrain updates to previous parameters to preserve learnt knowledge. In many popular approaches, including Elastic Weight Consolidation \citep{Kirkpatrick2017}, Gradient Episodic Memory \citep{GEM} and Synaptic Intelligence \citep{Zenke2017}, this is achieved by applying a loss to penalise changes to parameters considered to be important. Learning without Forgetting \citep{lwf} is another popular regularisation technique which records a model's output on new data and uses this as a pseudo-label to preserve the model's capabilities.

\textbf{Replay-based approaches} involve saving and retraining on previously seen data. Generally, it is assumed that only a small subset of training samples can be stored, in which case the main challenge is in selecting the best samples to store \citep{reservoir, GEM}. Other works in this area do not explicitly store the samples, but instead train a network to recreate them \citep{dgr}. Replay is commonly incorporated into continual learning algorithms, even if the works themselves are focused on other aspects \citep{zhu2022tame, hihn2023hierarchically, HCL}.

Continual learning scenarios are often categorised as either domain incremental or class incremental \citep{wang2024survey}, although other categories exist. In the domain incremental setting, the inputs for each task come from a different distribution and the output space is shared. In the class incremental setting, the inputs come from the same distribution and the output space is disjoint.

\subsection{Online Continual Learning Methods} \label{subsec:ocl}

In Online Continual Learning, also known as Task-Free or Task-Agnostic Continual Learning, task identities are not provided during training nor testing.\footnote{Online Continual Learning may also refer to scenarios where each training sample is only shown once, i.e.\ one epoch per task \citep{wang2024survey}. We do not consider this case and only focus on learning without task identities.} We provide an overview of some of the main approaches proposed in this area, which we will also use to compare with our method.

Bayesian Gradient Descent (BGD) is an early work in the Online Continual Learning setting \citep{BGD}. The authors build on previous work in online variational Bayesian learning to propose a closed-form update rule for the mean and variance of each parameter. Catastrophic forgetting is naturally mitigated as the posterior distribution of parameters, which captures all currently learnt knowledge, is used as the prior in the next training step.

\citet{CNDPM} formulate the Continual Learning problem as a stream of data from a mixture distribution and propose the Continual Neural Dirichlet Process Mixture (CN-DPM) algorithm where a set of experts, each containing a discriminative and a generative component, models the overall mixture distribution. In CN-DPM, new experts are added in a principled manner using Sequential Variational Approximation \citep{SVA}.

The Hybrid generative-discriminative approach to Continual Learning for classification (HCL) \citep{HCL} jointly models the distribution of the input data and labels for each task using a normalising flow. Three model statistics are tracked for each existing task and a task switch is detected if all three statistics lie outside the normal range. To prevent catastrophic forgetting, the authors propose using generative replay as well as a novel functional regularisation loss to enforce similar mappings to a snapshot taken on task switch.

In Task-Agnostic continual learning using Multiple Experts (TAME) \citep{zhu2022tame}, a set of experts is maintained where each is associated with a training task. Task switches are detected using a statistically significant deviation in the training loss, at which point a new expert is created. Each expert also maintains a small buffer of training samples, which are used to prune experts as well as train a selector network to choose the correct expert during test time.

Hierarchical Variational Continual Learning (HVCL) \citep{hihn2023hierarchically} extends VCL \citep{VCL} to the task-agnostic setting using the Sparsely Gated Mixture-of-Experts layer \citep{MoE}. Dynamic path selection through the HVCL model effectively produces distinct experts to solve different tasks. The authors introduce novel diversity objectives to better allocate experts to tasks, thus mitigating catastrophic forgetting.

Sparse neural Networks for Continual Learning (SNCL) \citep{SNCL} is a method that combines sparse networks, where unused parameters are reserved for future tasks, with a novel experience replay approach where the intermediate network activations are stored in addition to the sample. The sparsity is enforced by a variational Bayesian prior applied to the activation of each neuron. The proposed experience replay approach, named Full Experience Replay, allows for the priors to be more effectively optimised.

Finally, in Dynamically Expandable Representation (DER) \citep{DER}, a super-feature extractor is repeatedly expanded to accommodate new data. Catastrophic forgetting is mitigated as the older parameters are frozen, whilst pruning reduces redundancy in the added parameters. As the super-feature extractor is expanded, the classifier is continually finetuned using experience replay.

\section{\uppercase{Online Continual Learning}}
\label{sec:OCL}
In the Online Continual Learning paradigm, training data $D_{train} = \{ ( x_{i}, y_{i})\}_{i=1}^{N}$ in the form of input-output pairs arrives sequentially such that when $( x_{i}, y_{i})$ is seen, $\{ ( x_{j}, y_{j})\}_{j=1}^{i-1}$ cannot be accessed. The goal is to learn a model $f$ of inputs to outputs with no information given about the data generating process. Following training, the model is judged on testing data $D_{test} = \{ ( x_{i}, y_{i})\}_{i=1}^{M}$ by comparing the expected ($y_{i}$) and model ($f(x_{i})$) outputs.

We follow the common assumption \citep{hihn2023hierarchically, HCL} that there exists a set of distributions (tasks) from which the data is generated and that they arrive in a sequence (possibly with repetitions) with a hard boundary between each task. Formally, there exists a constant $C$ such that if $(x_{i}, y_{i})$ belongs to task $m$ and $(x_{i+1}, y_{i+1})$ belongs to task $m+1$, then $\{ ( x_{j}, y_{j})\}_{j=i+1}^{i+C}$ belongs to task $m+1$.

At a hard task boundary a large and sustained increase in the training loss can be expected. Task-Agnostic continual learning using Multiple Experts (TAME) \citep{zhu2022tame} (cf.\ Section~\ref{subsec:ocl}) is a method for detecting these boundaries by tracking the loss of a single active expert on incoming samples. A threshold for acceptable losses is calculated using the mean and standard deviation of the loss over a moving window. When the incoming loss, which is smoothed using an \emph{exponentially weighted moving average\/} (EWMA), exceeds this threshold, a task switch is detected. The active expert is then switched to the correct expert or to a new expert if the incoming task is unseen.

In this section, we propose a method for Online Continual Learning that, like TAME, leverages statistically significant increases in the training loss to detect task switches. First we describe our approach to task switch detection (Section~\ref{subsec:detecting}). Then we show how this can be applied to extend the Expert Gate algorithm \citep{ExpertGate} to produce our Gated Experts (GE) algorithm (Section~\ref{subsec:GE}). We also provide a comparison between TAME and GE (Section~\ref{subsec:TAME}).

\subsection{Detecting Task Switches in Online Continual Learning}
\label{subsec:detecting}

Given an expert and an input sample, we must determine if the sample belongs to the same task on which the expert was trained. To do so, for each expert we maintain an EWMA of the value and deviation of the training loss. Given $n$ training losses $L$, the mean $\mu$ and standard deviation $\sigma$ are defined as follows:
\begin{equation*}
\begin{aligned}
    \mu_{1} & \gets L_{1} \\
    \mu_{n} & \gets \alpha\,\mu_{n-1} + (1-\alpha)L_{n}, n > 1\\
    \sigma_{1} & \gets 0\\
    \sigma_{2} & \gets L_{2} - L_{1}\\
    \sigma_{n} & \gets \alpha\,\sigma_{n-1} + (1-\alpha)(L_{n} - \mu_{n-1}), n > 2\\
\end{aligned}
\end{equation*}
$\alpha$ is the smoothing factor for the EWMA and should be set relatively high (we use $\alpha=0.9$) to reduce the impact of noise in the training loss. The mean and standard deviation of the training loss are combined with a threshold hyperparameter $\epsilon$ to produce an \emph{upper bound on the training loss\/}:
\begin{equation*}
    \mu + \epsilon\sigma
\end{equation*}
If the loss of the given input sample is above this upper bound, we assume one of three possibilities: 
\begin{enumerate}
    \item The sample belongs to the same task as the expert but is an outlier.
    \item The sample belongs to another task.
    \item The sample belongs to the same task as the expert but instability in the training process has temporarily increased the loss.
\end{enumerate}
To determine which of these is correct, we set aside the input sample in a buffer (the \emph{high-loss buffer\/}) and await further samples. If the training loss immediately returns below the threshold, the first scenario has occurred as the other two can be ruled out.

It is clear that the second scenario will result in a consistently high training loss, and we also expect to see this trend in the third scenario; if training instability has left the expert in a state where it produces high training losses, setting aside the input samples prevents the expert from exiting such a state. Thus, if a consistently high training loss is observed, we must determine if the second or third scenario has occurred.

The second and third scenarios can be differentiated with the aid of a small buffer of samples on which the expert was trained. Samples from this buffer are compared to the high-loss samples in a statistical test to determine if they belong to the same distribution. First, the training loss is calculated for all samples. We hypothesise that the loss is normally distributed because in practice, each loss is the mean over a batch of training examples and as a result the Central Limit Theorem can be applied. Thus, we apply a simple Z-test\/: Let $\mu_{1}$ and $\sigma$ be the mean and standard deviation of the $n$ losses from the samples in the expert buffer, and $\mu_{2}$ be the mean of the losses from the high-loss buffer, then the standard error $SE$ and Z-score $ZS$ are calculated as:
\begin{equation*}
\begin{aligned}
    SE & = \sigma/\sqrt{n} \\
    ZS & = |\mu_{2} - \mu_{1}| / SE \\
\end{aligned}
\end{equation*}
If $ZS$ is above a threshold hyperparameter $\epsilon_{review}$, then we assume scenario 2 has occurred, and scenario 3 otherwise.

\subsection{Gated Experts}
\label{subsec:GE}

\begin{algorithm}[h]
 \footnotesize
 \DontPrintSemicolon
 \caption{Pseudocode for the GE algorithm}\label{alg:ge}
 \KwData{$(x_{i},y_{i})$ = the incoming training sample \\
    E = the set of existing experts \\
    E$_{new}$ = the set of newly created experts \\
    R = a buffer to store recent samples
 }
 \BlankLine
 append $(x_{i},y_{i})$ to R\;
 e$_{best}$ $\gets$ \textup{GE\_forward(E, $(x_{i},y_{i})$)} \tcp*[h]{returns the expert with lowest autoencoding loss}\;
 \eIf{\textup{Loss(e$_{best}$, $(x_{i},y_{i})$) $>$ e$_{best}$.threshold}}{
    \eIf{\textup{there exists e$_{new}$ in E$_{new}$ such that Loss(e$_{new}$, $(x_{i},y_{i})$) $\le$ e$_{new}$.threshold}}
    {
        train e$_{new}$ on $(x_{i},y_{i})$\;
        promote e$_{new}$ if the conditions are met\;
    }
    {
        mark $(x_{i},y_{i})$ as a high-loss sample\;
    }
 }{
    train e$_{best}$ on $(x_{i},y_{i})$\;
 }
 \BlankLine
 \tcp{Process outliers}
 \If{\textup{R is not full}}
 {
    return\;
 }
 $(x_{last},y_{last})$ $\gets$ the oldest sample in R\;
 \If{\textup{$(x_{last},y_{last})$ is marked as high-loss}}
 {
    e$_{last}$ $\gets$ the expert that was trained on the previous sample to $(x_{last},y_{last})$\;
    train e$_{last}$ on $(x_{last},y_{last})$\;
 }
 remove $(x_{last},y_{last})$ from R\;
 $(x_{last},y_{last})$ $\gets$ the new oldest sample in R\;
 \BlankLine
 \tcp{Detect task switches}
 \If{\textup{every remaining entry in R is a high-loss sample}}
 {
    e$_{last}$ $\gets$ the expert with the lowest autoencoding loss on $(x_{last},y_{last})$\;
    test the loss of e$_{last}$ on the samples in R and the samples in e$_{last}$.replay\_buffer\;
    \eIf{\textup{the losses are statistically different}}
    {
        create a new expert e$_{new}$\;
        train e$_{new}$ on the samples in R\;
    }
    {
        train e$_{last}$ on the samples in R\;
    }
    remove all samples from r\;
 }

\end{algorithm}

Detecting switches in the input task allows for the Expert Gate algorithm \citep{ExpertGate} to be extended to the Online Continual Learning setting. We name this extension \emph{Gated Experts\/} (GE) and provide the pseudocode in Algorithm \ref{alg:ge}. In GE, a set of experts (initially one) is maintained with each expert corresponding to a single task out of all tasks seen so far. Associated with each expert is an autoencoder that is trained on the same data. As each training sample arrives, it is forwarded to the most relevant expert defined by the lowest autoencoding loss. Thus, task switches between existing tasks are handled. When a task switch is detected, we can assume that the incoming task is new and thus create a new expert.

When a new expert is created, it is likely that future samples from the same task will not be correctly forwarded to said expert until the associated autoencoder is adequately trained. GE addresses this problem by maintaining a separate set of newly created experts. When a sample is forwarded to an expert and found to not belong to said expert, we check if it belongs to a newly created expert before setting it aside. Thus, the following events will occur when samples from a new task arrive:
\begin{enumerate}
    \item The samples are forwarded to one or more existing experts, found to have a high loss and thus set aside in the high-loss buffer.
    \item A task switch is detected due to a consistently high loss, and a new expert is created and trained on the samples in the high-loss buffer.
    \item Further samples from the new task arrive and are forwarded to the same experts as in Step 1.
    \item The samples are redirected to the new expert rather than sent to the high-loss buffer.
\end{enumerate}

During the final step, we also track whether the loss of the sample on the new expert was greater or lower than on the existing expert. When the loss on the new expert is consistently lower, then the new expert can be promoted: the expert is removed from the set of newly created experts and treated as a regular expert. The promotion condition is met when the proportion of lower losses in the last $promotion\_window$ samples exceeds $\epsilon_{promotion}$. In GE, we set $promotion\_window = 50$ and $\epsilon_{promotion} = 0.5$.

A relatively low value of $\epsilon_{promotion}$ is possible as GE does not require perfect accuracy when forwarding samples. Under the assumption of hard task boundaries, adjacent samples are highly likely to belong to the same task. When a sample is forwarded to the wrong expert, a temporary high-loss is observed similar to an outlier. In both cases, the sample will be assigned to the same expert as the previous sample.

\begin{algorithm}[h]
 \small
 \DontPrintSemicolon
 \caption{HGE\_forward - In HGE, the most relevant expert for a given sample is found by starting at the root and repeatedly choosing the best child.}\label{alg:hge_forward}
 \KwData{$(x_{i},y_{i})$ = the incoming sample \\
    E = the tree of existing experts \\
 }
 \KwResult{e$_{best}$ = the expert with the lowest autoencoding loss on $(x_{i},y_{i})$
 }
 \BlankLine
 n $\gets$ the root node of E\;
 \While{True}
 {
    \If{\textup{n has no children}}
    {
        return\;
    }
    c $\gets$ the node in n.children with the lowest AE\_Loss(c.expert, $(x_{i},y_{i})$)\;
    \eIf{\textup{e$_{best}$ does not yet exist or AE\_Loss(c.expert, $(x_{i},y_{i})$) $<$ AE\_Loss(e$_{best}$, $(x_{i},y_{i})$)}}
    {
        e$_{best}$ $\gets$ c.expert\;
    }
    {
        return\;
    }
 }
\end{algorithm}

\subsection{Comparison to TAME}
\label{subsec:TAME}

The main advantage of GE over TAME is in the high-loss buffer. Performing a statistical test before detecting a task switch greatly reduces the false-positive rate, as we show experimentally in Section~\ref{subsec:task_switch}. In addition, in TAME one or more samples may be wrongly trained on the active expert before the EWMA of the loss exceeds the acceptable threshold. This is prevented in GE as samples over the threshold are immediately set aside.

Another difference is in the lack of an active expert in GE. By assigning each sample to the expert with the lowest autoencoding loss, the algorithm is simplified as task switches between existing tasks are a non-factor. In practice, the last-used expert could be tracked to improve the efficiency of GE. If an incoming sample has a loss within the threshold of the last-used expert, then we can assign it to said expert without calculating all of the autoencoding losses.

\section{\uppercase{Hierarchically Gated Experts}}
\label{sec:HGE}

In GE the autoencoding losses are used as a measure of the suitability of an expert to a given sample. We hypothesise the existence of relationships between experts such that \emph{the suitability of an expert can be estimated from the autoencoding losses of its peers\/}. For example, if experts A and B are both trained on tasks in the same domain then we can expect that they will produce a low autoencoding loss for any sample from said domain. Thus, if expert A produces a high autoencoding loss then we can assume that the sample originates from a different domain, and that expert B will likewise be unsuitable.

We propose the idea of organising experts \emph{hierarchically\/} to exploit these relationships and improve the speed with which an expert can be selected for a given sample. Our algorithm, named \emph{Hierarchically Gated Experts\/} (HGE), is an extension to GE where the set of existing experts is instead a tree with each node (except the root) corresponding to a specific expert. Given a sample, we traverse the tree by following the child with the lowest autoencoding loss at each step. We stop when a leaf node is reached or when the current node has a lower autoencoding loss than each of its children. The final node is then considered to correspond to the expert with the overall lowest autoencoding loss. Thus, an expert can be selected without calculating the autoencoding loss for all experts. The pseudocode for the traversal is provided in Algorithm~\ref{alg:hge_forward}.

\begin{algorithm}[h]
 \small
 \DontPrintSemicolon
 \caption{Expert Promotion in HGE}\label{alg:hge_promotion}
 \KwData{E = the tree of existing experts \\
    e$_{new}$ = the expert to be promoted \\
    paths = a list of the unique paths taken by the samples on which e$_{new}$ was trained. Each path also contains a count of the number of times it was taken.
 }
 \BlankLine
 e$_{new}$\_node $\gets$ a new node with e$_{new}$\_node.expert = e$_{new}$ and no children\;
 \If{\textup{there is only one expert in E}}
 {
    add e$_{new}$\_node as a child of the root node of E\;
    return\;
 }
 \BlankLine
 \tcp{Insert expert into tree}
 sort paths by count in decreasing order\;
 \For{\textup{p in paths}}
 {
    sum $\gets$ the cumulative sum of counts up to and including p\;
    total $\gets$ the sum of all counts in paths\;
    \If{\textup{sum $>$ promotion\_path\_threshold * total}}
    {
        remove all remaining members of paths\;
    }
 }
 parent $\gets$ the LCA of all remaining p in paths\;
 add e$_{new}$\_node as a child of parent\;
 \BlankLine
 \tcp{Add backward connections}
 possible\_overwrites $\gets$ \{\}\;
 \For{\textup{each descendant n of parent excluding e$_{new}$\_node}}
 {
    add n.expert to possible\_overwrites\;
 }
 \For{\textup{e$_{overwrite}$ in possible\_overwrites}}
 {
    \If{\textup{there exists $(x_{i},y_{i})$ in e$_{overwrite}$.replay\_buffer where HGE\_forward(E, $(x_{i},y_{i})$) = e$_{new}$}}
    {
        add a node n as a child of e$_{new}$\_node with n.expert = e$_{overwrite}$\;
    }
 }
\end{algorithm}

The tree of experts is built iteratively. As each expert is promoted, it is added to the tree such that samples from the corresponding task will be correctly assigned to the expert. This can be trivially accomplished by adding the expert as a child of the root, so that its autoencoding loss will always be calculated. However, such a strategy will produce a completely flat tree and be equivalent to the GE algorithm. Thus, we instead aim to insert the expert as deeply as possible while maintaining assignment accuracy. As a newly created expert is trained, we track the traversal paths taken by every sample the expert is given. When the expert is promoted, it is inserted as a child of the lowest common ancestor (LCA) of all traversal paths. Outlier paths, which are only taken a few times, are excluded according to the $promotion\_path\_threshold$ hyperparameter. The pseudocode for expert promotion is provided in Algorithm \ref{alg:hge_promotion}.

\begin{figure}
    \centering
    \includesvg[width=.24\textwidth]{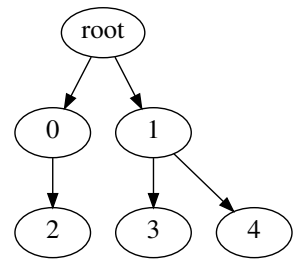}\hfill
    \caption{An example tree generated by HGE. Each node corresponds to a different expert.}
    \label{fig:hge_mnist}
\end{figure}

When a new expert is added to the tree, it is possible that it will mask some of the existing experts, causing catastrophic forgetting. With reference to Figure~\ref{fig:hge_mnist}, expert 0 will mask expert 3 if, on samples from task 3, expert 0 has a lower autoencoding loss than expert 1. When an expert is inserted, all descendants of siblings can potentially be masked. To mitigate this, the replay buffer of each potentially masked expert is used to estimate if masking will occur. If any replay samples are incorrectly assigned to the new expert, then a new node corresponding to the masked expert is added as a child of the new expert. We choose to create a new node rather than add a connection to the existing node so that we do not unnecessarily add the descendants of the masked expert.

The hierarchical organisation of experts complicates the promotion process. In GE, there is a wide range of possible promotion times as the algorithm is able to handle samples being forwarded to the wrong expert. In HGE, however, it is much easier for the promotion of an expert to be performed too early. When an expert is promoted in HGE, it is added to the tree in a way that exploits the relationship between the autoencoders. As the expert continues to train, these relationships may change and the expert may begin to mask others, causing catastrophic forgetting. Thus, a higher $\epsilon_{promotion}$ is preferred. On the other hand, a promotion threshold that is too high may never be met. 

In our experiments, we set a separate $\epsilon_{promotion}$ for each scenario roughly equal to the accuracy achieved at the end of training. This approach works reasonably well, but it does break the Online Continual Learning paradigm as we are using task-specific information. The promotion time is part of a larger problem of dealing with changing relationships between experts, which we leave as a problem for future research.

\section{\uppercase{Experiments}}
\label{sec:experiments}
In this section, we experimentally evaluate our approach to task switch detection in the Online Continual Learning setting. We also perform controlled experiments with GE and HGE to determine the impact of hierarchical expert organisation on the gating accuracy. Finally, we validate GE and HGE on Online Continual Learning benchmarks.

We use widely known and commonly adopted Continual Learning scenarios in our experiments:

\textbf{Permuted MNIST (PMNIST)}. The MNIST dataset \citep{MNIST} is used as the first task in this scenario. Each subsequent task reuses MNIST, but with the pixels in each image randomly permuted. The permutations are kept consistent within each task. There are 20 tasks in total. 

\textbf{Split MNIST (SMNIST)}. The MNIST dataset is split into 5 tasks, each containing 2 classes.

\textbf{Split CIFAR-10 (CIF10)}. The CIFAR-10 dataset \citep{CIFAR} is split into 5 tasks, each containing 2 classes.

\textbf{Split CIFAR-100 (CIF100)}. The CIFAR-100 dataset is split into either 10 or 20 tasks, each with the same number of classes.

\textbf{Split Tiny Imagenet (ImgNet)}. The Tiny Image\-net dataset is a subset of Imagenet \citep{imagenet} containing images from 200 classes. We randomly select 100 classes which are then evenly split into either 10 or 20 tasks.

We also use scenarios that are both class-incre\-men\-tal and domain-incremental. Each scenario involves two datasets where the classes are split evenly into tasks. The tasks are presented in an alternating fashion; the first task from the first dataset, then the first task from the second dataset, then the second task from the first dataset, and so on. The scenarios are:

\textbf{MNIST and Kuzushiji-MNIST (MNIST-KMNIST)}. We split the MNIST dataset and Kuzushiji-MNIST (KMNIST) dataset \citep{kmnist} into 5 tasks each, for a total of 10 tasks.

\textbf{MNIST and CIFAR-10 (MNIST-CIF10)}. Similar to the previous scenario, we split each dataset into 5 tasks each.

\textbf{CIFAR-10 and Inverse CIFAR-10 (CIF10-INV)}. We split each dataset into 5 tasks. To produce the Inverse CIFAR-10 dataset, we invert the colour of every pixel. The dataset is otherwise identical to CIFAR-10.

\textbf{CIFAR-100 and Tiny Imagenet (CIF-ImgNet)}. Each dataset is split into 10 tasks. Only 100 randomly selected classes are used from the Tiny Imagenet dataset.

We use two variants of experts in our experiments. In the PMNIST, SMNIST and MNIST-KMNIST scenarios, both the base model and the autoencoder are multi-layer perceptrons (MLP). Otherwise, we use Resnet-18 \citep{resnet} as the base model and a convolutional autoencoder. We provide the full details of the dataset preprocessing, models used and hyperparameters in the Appendix.

\subsection{Task Switch Detection}
\label{subsec:task_switch}
\begin{table}[h]
\caption{The number of task switch detection errors per task, categorised as false positives (FP) or false negatives (FN). Cases with runaway expert creation are marked DNF (Did Not Finish). CIFAR (all) and Imagenet (all) refer to all scenarios tested purely on these datasets.}
\centering
{\scriptsize
\begin{tabular}{|c|c|c|c|c|c|c|} 
 \hline
  & \multicolumn{2}{c|}{GE} & \multicolumn{2}{c|}{GE w/o review} & \multicolumn{2}{c|}{TAME} \\
 \hline
 Scenario & FP & FN & FP & FN & FP & FN \\
 \hline
 PMNIST & 0 & 0 & 0 & 0 & 0 & 0 \\
 SMNIST & 0 & 0 & 0 & 0 & 0.04 & 0.04 \\
 CIFAR (all) & 0 & 0 & DNF & DNF & DNF & DNF \\
 Imagenet (all) & 0 & 0 & DNF & DNF & DNF & DNF \\
 MNIST-KMIST & 0 & 0.04 & 0 & 0 & 0 & 0 \\
 MNIST-CIF10 & 0 & 0 & DNF & DNF & DNF & DNF \\
 CIF10-INV & 0 & 0 & DNF & DNF & DNF & DNF \\
 CIF-ImgNet & 0 & 0 & DNF & DNF & DNF & DNF \\
 \hline
\end{tabular}}
\label{tab:task_switch}
\end{table}

\begin{table*}[h]
\small
\caption{The accuracy and efficiency of different approaches to organising the expert tree. Acc refers to the percentage of batches which were assigned to the correct expert, whilst Exp is the average number of experts queried.}
\centering
\begin{tabular}{|c|c|c|c|c|c|c|} 
 \hline
  & \multicolumn{2}{c|}{GE} & \multicolumn{2}{c|}{HGE} & \multicolumn{2}{c|}{Upper} \\
 \hline
 Scenario & Acc & Exp & Acc & Exp & Acc & Exp \\
 \hline
 PMNIST & $100.0\pm0.0$ & $20$ & $99.76\pm0.11$ & $11.25\pm1.02$ & $99.7\pm0.19$ & $8.05\pm0.13$ \\
 SMNIST & $100.0\pm0.0$ & $5$ & $98.26\pm1.4$ & $3.83\pm0.21$ & $99.75\pm0.56$ & $3.56\pm0.09$ \\
 CIF10 & $95.25\pm5.11$ & $5$ & $95.0\pm5.66$ & $4.17\pm0.55$ & $95.0\pm5.66$ & $3.85\pm0.38$ \\
 CIF100(10) & $77.25\pm2.05$ & $10$ & $76.75\pm1.9$ & $9.49\pm0.85$ & $75.0\pm4.59$ & $8.02\pm0.33$ \\
 CIF100(20)  & $83.25\pm8.08$ & $20$ & $82.75\pm7.52$ & $17.55\pm1.43$ & $84.0\pm8.59$ & $12.84\pm0.74$ \\
 ImgNet(10) & $61.0\pm13.3$ & $10$ & $61.0\pm13.3$ & $9.7\pm0.42$ & $59.0\pm12.57$ & $8.47\pm0.5$ \\
 ImgNet(20) & $73.5\pm5.76$ & $20$ & $73.5\pm5.76$ & $19.02\pm0.93$ & $73.5\pm2.85$ & $15.03\pm0.6$ \\
 MNIST-KMIST & $95.25\pm5.94$ & $10$ & $95.25\pm5.94$ & $6.42\pm0.57$ & $95.01\pm5.72$ & $5.13\pm0.09$ \\
 MNIST-CIF10 & $98.88\pm0.52$ & $10$ & $98.75\pm0.77$ & $5.29\pm0.1$ & $98.4\pm1.01$ & $4.92\pm0.16$ \\
 CIF10-INV & $98.25\pm0.93$ & $10$ & $97.88\pm0.71$ & $6.62\pm0.99$ & $98.0\pm0.93$ & $5.31\pm0.19$ \\
 CIF-ImgNet & $68.88\pm1.49$ & $20$ & $68.88\pm1.49$ & $19.06\pm0.81$ & $68.38\pm1.44$ & $15.6\pm0.65$ \\
 \hline
\end{tabular}
\label{tab:tree_efficiency}
\end{table*}

We first test the effectiveness of our approach to task switch detection. In each scenario, we observe the creation of new experts and record cases where an abnormal number of experts is created for a task. Each additional expert is recorded as a false positive, whereas the case where no expert is created is recorded as a false negative. When runaway expert creation is observed (more than five per task), we terminate the experiment early. We compare the GE algorithm with TAME and also perform an ablation test (GE w/o review) where we apply GE but do not perform a statistical test before creating a new expert.

Our results are summarised in Table \ref{tab:task_switch}. The low error rate of GE across all scenarios demonstrates the efficacy of our approach to task switch detection. The TAME and ablation test results reflect the nature of the base models used in our experiments. The 3-layer MLP is relatively stable during training, and as a result we see very few errors in the pure MNIST scenarios. On the other hand, our Resnet-18 model has high instability, which causes runaway expert creation. These results show the importance of the statistical test in differentiating between spikes in the loss due to training instability and spikes due to data from a new task.

We observed a small number of false negatives for GE in the MNIST-KMNIST scenario. Upon analysis, we found that the algorithm was able to re-use an existing expert with no loss in accuracy.

In our experiments, we observed much higher z-scores than expected; over 20,000 during task switches and up to 10 during some spikes in the training loss. Thus, our assumption that the training loss is normally distributed does not hold well in practice. However, we still saw a clear separation between the z-scores in both situations, and tuning the z-score threshold hyperparameter $\epsilon_{review}$ was not necessary. We set $\epsilon_{review}$ to a relatively high value and found this worked for all scenarios.

\subsection{Hierarchical Organisation}
\label{subsec:hierarchical_organisation}

We aim to determine the impact of hierarchical organisation on the accuracy and efficiency of the expert selection process. We control for other factors in Online Continual Learning, such as training samples being assigned to the wrong expert, by instead training a separate expert on each task. We only measure the accuracy with which a sample is assigned to the correct expert and do not consider the classification accuracy. We measure the cost of a tree as the number of experts queried for a given sample.
After all experts are trained, we apply three approaches to expert organisation:

\textbf{GE}. The experts are organised in a fully flat tree.

\textbf{HGE}. In the same manner as the HGE algorithm, a tree of experts is built by incrementally adding experts according to task order. As each expert is trained separately, we do not have any traversal paths to track. Instead, when an expert is added to the tree the traversal paths are calculated for all training samples in the corresponding task.

\textbf{Upper}. The HGE algorithm is run a large number of times (1000) with a randomised task order, and the accuracy and cost of each generated tree is recorded. We then select the tree with the lowest cost and an accuracy at least as high (approximately) as the GE tree. This approach estimates the best possible tree given the trained experts.

Our results are shown in Table~\ref{tab:tree_efficiency}. Both the HGE and Upper approaches are able to produce trees with close to optimal accuracies. In general, the cost reduction with HGE is dependent on the expert selection accuracy, with a markedly lower cost in scenarios where the accuracy is high. This trend is also observed with the Upper strategy, but to a lesser extent as there is a decent cost reduction even when the selection accuracy is low.

\begin{table*}[h]
\small
\caption{A statistical analysis of the accuracy and query costs of all trees generated while finding an upper bound. PCC and SCC are respectively the Pearson and Spearman correlation coefficients, whilst MAD is the median absolute deviation. Each metric is computed over all trees generated for a set of experts, then averaged across every set.}
\centering
\begin{tabular}{|c|c|c|c|c|c|c|c|c|c|c|c|c|} 
 \hline
  &  & \multicolumn{5}{c|}{Acc} \\
 \hline
 Scenario & PCC & Mean & STD & Median & IQR & MAD \\
 \hline
 PMNIST & $0.13\pm0.06$ & $99.57\pm0.04$ & $0.45\pm0.08$ & $99.71\pm0.03$ & $0.39\pm0.07$ & $0.19\pm0.0$ \\
 SMNIST & $0.3\pm0.19$ & $99.55\pm0.48$ & $0.7\pm0.6$ & $100.0\pm0.0$ & $0.72\pm1.06$ & $0.0\pm0.0$ \\
 CIF10 & $0.02\pm0.03$ & $95.08\pm5.46$ & $0.28\pm0.41$ & $95.12\pm5.38$ & $0.25\pm0.56$ & $0.12\pm0.28$ \\
 CIF100(10) & $0.55\pm0.19$ & $77.03\pm1.98$ & $0.53\pm0.13$ & $77.25\pm2.05$ & $0.25\pm0.56$ & $0.0\pm0.0$ \\
 CIF100(20) & $0.13\pm0.18$ & $83.07\pm7.93$ & $0.86\pm0.09$ & $83.25\pm8.08$ & $0.25\pm0.56$ & $0.0\pm0.0$ \\
 ImgNet 10) & $0.11\pm0.29$ & $60.97\pm13.28$ & $0.39\pm0.2$ & $61.0\pm13.3$ & $0.0\pm0.0$ & $0.0\pm0.0$ \\
 ImgNet(20) & $0.18\pm0.22$ & $73.42\pm5.6$ & $1.06\pm0.24$ & $73.5\pm5.76$ & $0.0\pm0.0$ & $0.0\pm0.0$ \\
 MNIST-KMIST & $0.09\pm0.12$ & $95.13\pm5.91$ & $0.36\pm0.1$ & $95.25\pm5.94$ & $0.12\pm0.28$ & $0.0\pm0.0$ \\
 MNIST-CIF10 & $0.01\pm0.08$ & $98.78\pm0.57$ & $0.21\pm0.26$ & $98.88\pm0.52$ & $0.12\pm0.28$ & $0.0\pm0.0$ \\
 CIF10-INV & $0.14\pm0.12$ & $98.03\pm0.84$ & $0.52\pm0.17$ & $98.25\pm0.93$ & $0.12\pm0.28$ & $0.12\pm0.28$ \\
 CIF-ImgNet & $0.31\pm0.12$ & $68.77\pm1.47$ & $0.44\pm0.14$ & $68.88\pm1.49$ & $0.0\pm0.0$ & $0.0\pm0.0$ \\
 \hline
  &  & \multicolumn{5}{c|}{Exp} \\
 \hline
 Scenario & SCC & Mean & STD & Median & IQR & MAD \\
 \hline
 PMNIST & $0.17\pm0.08$ & $11.32\pm0.32$ & $1.45\pm0.14$ & $11.16\pm0.34$ & $1.97\pm0.22$ & $0.97\pm0.1$ \\
 SMNIST & $0.36\pm0.21$ & $4.08\pm0.09$ & $0.32\pm0.03$ & $4.07\pm0.15$ & $0.41\pm0.02$ & $0.26\pm0.07$ \\
 CIF10 & $0.03\pm0.07$ & $4.29\pm0.36$ & $0.31\pm0.05$ & $4.25\pm0.33$ & $0.55\pm0.17$ & $0.22\pm0.12$ \\
 CIF100(10) & $0.54\pm0.21$ & $9.72\pm0.14$ & $0.39\pm0.09$ & $9.87\pm0.29$ & $0.45\pm0.3$ & $0.1\pm0.23$ \\
 CIF100(20) & $0.12\pm0.2$ & $17.56\pm0.89$ & $1.37\pm0.11$ & $17.65\pm0.97$ & $1.78\pm0.18$ & $0.88\pm0.12$ \\
 ImgNet 10) & $0.05\pm0.31$ & $9.86\pm0.1$ & $0.26\pm0.12$ & $10.0\pm0.0$ & $0.13\pm0.29$ & $0.0\pm0.0$ \\
 ImgNet(20) & $0.15\pm0.21$ & $19.07\pm0.33$ & $0.91\pm0.18$ & $19.19\pm0.31$ & $1.25\pm0.35$ & $0.64\pm0.12$ \\
 MNIST-KMIST & $0.1\pm0.12$ & $6.51\pm0.18$ & $0.64\pm0.09$ & $6.46\pm0.2$ & $0.87\pm0.19$ & $0.45\pm0.07$ \\
 MNIST-CIF10 & $0.02\pm0.1$ & $5.91\pm0.13$ & $0.72\pm0.04$ & $5.62\pm0.13$ & $0.94\pm0.12$ & $0.34\pm0.05$ \\
 CIF10-INV & $0.14\pm0.13$ & $7.25\pm0.28$ & $0.85\pm0.13$ & $7.22\pm0.32$ & $1.18\pm0.18$ & $0.57\pm0.08$ \\
 CIF-ImgNet & $0.27\pm0.13$ & $19.37\pm0.14$ & $0.73\pm0.09$ & $19.62\pm0.35$ & $1.02\pm0.27$ & $0.38\pm0.35$ \\
 \hline
\end{tabular}
\label{tab:HGER_metrics}
\end{table*}

We also provide a statistical analysis of the accuracies and costs of the trees generated during the Upper strategy. We analyse the spread of the accuracy and cost as well as the correlation between the two. The results are provided in Table~\ref{tab:HGER_metrics}. We observe a low but positive level of correlation across the board. We believe that deeper trees are able to have a lower cost, but may also suffer a lower accuracy as deeper nodes have a higher chance of being masked.

We observe a very low level of spread in the accuracies of the trees. The spread of the costs is higher but still relatively low. Thus, we conclude that HGE is not overly affected by the task order and is generally able to organise the experts with close to the optimal possible efficiency and cost.

\begin{figure}[ht]
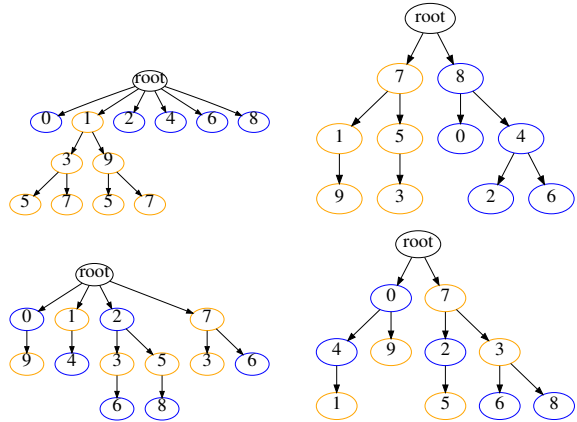

    \scriptsize
    \centering
    \includesvg[width=.22\textwidth]{HGE_CIFAR_INVCIFAR2_split.json.svg}\hfill
    \includesvg[width=.22\textwidth]{HGER_CIFAR_INVCIFAR_split.json.svg}
    \\[\smallskipamount]
    \includesvg[width=.22\textwidth]{HGE_MNIST_KMNIST_split.json.svg}\hfill
    \includesvg[width=.22\textwidth]{HGER_MNIST_KMNIST_split.json.svg}
    \caption{Examples of trees generated by HGE and Upper. Clockwise starting from the top-left: HGE on CIF10-INV, Upper on CIF10-INV, Upper on MNIST-KMNIST, HGE on MNIST-KMNIST. The nodes are coloured according to the domain.}
    \label{fig:hge_upper_comparison}
\end{figure}

A manual examination of the trees confirms these findings, as the HGE trees share many similarities with the Upper trees. Figure \ref{fig:hge_upper_comparison} showcases this; both HGE and Upper organise the experts according to domain in CIF10-INV, and in MNIST-KMNIST we see identical connections such as between experts 7 and 3 or 2 and 5.

\subsection{Continual Learning Benchmarks}
\label{subsec:benchmarks}
We evaluate the overall efficacy of GE and HGE by providing a benchmark of the accuracies achieved by these algorithms in an Online Continual Learning setting. We also provide an estimate of the accuracy that can be achieved given the base models used. The \textit{Separate} method creates a separate model for each task and uses task information to assign the samples, thus serving as an upper bound.

\begin{table*}[h]
\footnotesize
\caption{Accuracy on continual learning benchmarks (\%)}
\centering
\begin{tabular}{|c|c|c|c|c|c|c|c|c|} 
 \hline
 Method & PMNIST & SMNIST & CIF10 & CIF100(10) & CIF100(20) & ImgNet(10) & ImgNet(20) \\
 \hline
 Separate & $ 97.9 \pm 0.0 $ & $ 99.44 \pm 0.15 $ & $ 93.1 \pm 2.03 $ & $ 70.96 \pm 0.32 $ & $ 80.1 \pm 0.89 $ & $ 53.76 \pm 1.04 $ & $ 67.3 \pm 0.62 $ \\
 \hline
 GE & $ 97.9 \pm 0.07 $ & $ 99.54 \pm 0.09 $ & $ 89.06 \pm 4.05 $ & $ 51.96 \pm 6.31 $ & $ 65.62 \pm 5.76 $ & $ 31.02 \pm 5.3 $ & $ 49.26 \pm 3.17 $ \\
 \!\!Gate Acc\!\! & $ 100.0 \pm 0.0 $ & $ 100.0 \pm 0.0 $ & $ 96.78 \pm 3.36 $ & $ 73.02 \pm 8.41 $ & $ 81.04 \pm 7.09 $ & $ 59.0 \pm 9.45 $ & $ 71.5 \pm 5.76 $ \\
 Exp & 20 & 5 & 5 & 10 & 20 & 10 & 20 \\
 \hline
 HGE & $ 82.3 \pm 7.01 $ & $ 99.48 \pm 0.04 $ & $ 87.08 \pm 9.1 $ & $ 47.42 \pm 4.62 $ & $ 67.76 \pm 7.92 $ & $ 30.68 \pm 2.07 $ & $ 42.48 \pm 6.28 $ \\
 \!\!Gate Acc\!\! & $ 82.18 \pm 8.03 $ & $ 100.0 \pm 0.0 $ & $ 92.25 \pm 10.21 $ & $ 68.28 \pm 5.75 $ & $ 83.54 \pm 10.41 $ & $ 57.5 \pm 3.95 $ & $ 61.5 \pm 9.78 $ \\
 Exp & $ 8.75 \pm 0.97 $ & $ 4.37 \pm 0.49 $ & $ 4.26 \pm 0.61 $ & $ 10.0 \pm 0.0 $ & $ 17.79 \pm 1.23 $ & $ 9.91 \pm 0.19 $ & $ 18.91 \pm 0.54 $ \\
 \hline
 BGD & 79.15 & 19.00 & N/A & N/A & 3.77 & N/A & N/A \\
 HCL & N/A & 90.89 & 89.44 & 59.66 & N/A & N/A & N/A \\
 TAME & 87.32 & 98.63 & 91.32 & 61.06 & 62.39 & N/A & N/A \\
 HVCL & 97.47 & 98.60 & 81.00 & 37.20 & N/A & N/A & N/A \\
 \!\!\!CN-DPM\!\!\! & N/A & 93.23 & 46.98 & N/A & 20.10 & N/A & N/A \\
 SNCL & 92.93 & N/A & 90.41 & N/A & N/A & N/A & 39.83 \\
 DER & 91.66 & N/A & 83.81 & 74.64 & 73.98 & N/A & 36.73 \\
 HN & N/A & N/A & N/A & N/A & 73.6 & N/A & 39.9 \\
 \hline
\end{tabular}
\label{tab:benchmark}
\end{table*}

\begin{table*}[h]
\small
\caption{Accuracy on hybrid scenarios (\%)}
\centering
\begin{tabular}{|c|c|c|c|c|} 
 \hline
 Method & MNIST-KNIST & MNIST-CIF10 & CIF10-INV & CIF-ImgNet \\
 \hline
 Separate & $ 99.54 \pm 0.05 $ & $ 96.56 \pm 0.62 $ & $ 93.4 \pm 1.26 $ & $ 63.5 \pm 0.71 $ \\
 \hline
 GE & $ 99.38 \pm 0.16 $ & $ 95.26 \pm 1.42 $ & $ 90.3 \pm 1.86 $ & $ 41.68 \pm 4.65 $ \\
 Gate Acc & $ 100.0 \pm 0.0 $ & $ 98.52 \pm 1.22 $ & $ 96.88 \pm 1.16 $ & $ 66.16 \pm 8.2 $ \\
 Exp & 10 & 10 & 10 & 20 \\
 \hline
 HGE & $ 94.18 \pm 8.33 $ & $ 88.62 \pm 7.65 $ & $ 85.84 \pm 8.09 $ & $ 37.6 \pm 3.74 $ \\
 Gate Acc & $ 93.16 \pm 8.45 $ & $ 91.76 \pm 7.51 $ & $ 91.9 \pm 8.01 $ & $ 59.9 \pm 7.23 $ \\
 Exp & $ 6.66 \pm 1.22 $ & $ 5.08 \pm 0.31 $ & $ 7.47 \pm 1.56 $ & $ 19.22 \pm 0.65 $ \\
 \hline
\end{tabular}
\label{tab:hybrid_benchmark}
\end{table*}

Our results are presented in Table \ref{tab:benchmark} and \ref{tab:hybrid_benchmark}. We measure accuracy as the average accuracy achieved over all tasks at the end of training. We also report the accuracy with which batches were assigned to the correct expert (Gate Acc). During training, we track the samples assigned to each expert and record an association between an expert and a task if the expert was trained on at least 10\% of the samples in the task. During testing, we then measure the percentage of samples assigned to the correct expert. Finally, we also report the average number of experts queried during testing.

For comparison, we also provide the accuracy achieved by all the other existing continual learning methods described in Section~\ref{subsec:ocl}.
The BGD results are taken from \citet{zhu2022tame} and part of the DER results from \citet{SNCL}. All other results are as reported in their corresponding papers cited in Section~\ref{subsec:ocl}.

Overall, GE is able to match or exceed the state-of-the-art in most benchmarks. Although the comparison with other techniques is not entirely fair due to some differences in experimental setup, the results still show that our algorithm is competitive with the compared techniques in Online Continual Learning.

One exception is the 10-task split CIFAR-100 and Tiny Imagenet scenarios, where there is a low gating accuracy. We hypothesise that the higher number of classes per task leads each autoencoder to learn a more general representation of the data rather than fitting specifically to the task. To test this, we exploit the additional labels in CIFAR-100, where each class is part of 1 of 20 superclasses. If we split by pairs of superclasses rather than randomly, the gating accuracy increases to 87.5\%, however the overall accuracy increases only to 52.65\% due to a lower classification accuracy between classes within a superclass.

In HGE, we observed a decreased accuracy across the board compared with GE. Whilst this is to be expected, we found larger drops in scenarios where the number of experts queried also distinctly dropped. Thus, we hypothesise that the main cause of the decreased accuracy is the changing relationships between experts as described in Section~\ref{sec:HGE}.

\section{\uppercase{Conclusion}}
\label{sec:conclusion}
In this paper, we proposed an alternative to TAME for task switch detection in the Online Continual Learning setting. We showed that our approach is better equipped to handle models with unstable training, and applied this approach to extend Expert Gate \citep{ExpertGate} to our online Gated Experts (GE) algorithm. Then, we proposed the novel HGE extension, where experts are organised hierarchically, as a method to improve the efficiency of GE and possibly other expert-based algorithms. We performed controlled experiments to determine that HGE is able to organise a set of experts in a manner that is close to optimal. Finally, we benchmarked GE and HGE on standard Continual Learning datasets and found GE to be competitive with the state-of-the-art.

The major problem of HGE is in the changing relationships between experts as they are trained, which causes samples to be assigned incorrectly. As a result, we observed a lower HGE accuracy in the Online Continual Learning setting. The changing relationships also make it difficult to determine the optimal time to promote a newly created expert. Future research could be conducted into methods for periodically updating the expert tree to accommodate the changing relationships.

HGE also suffers from poor efficiency when adding a new expert to the tree, as all descendants of siblings have to be checked for masking. More efficient approaches to building the expert tree may be possible. The use of autoencoders is also not strictly necessary and other methods for measuring expert suitability could be considered.

\bibliographystyle{apalike}
{\small
\bibliography{Example}}

\section*{\uppercase{Appendix}}
In this section we provide hyperparameters and other details crucial to recreating our results.

\textbf{Datasets}
In the PMNIST, SMNIST and MNIST-KMNIST scenarios, the 28x28 images are flattened into 784-dimensional vectors. In all other scenarios, all images are first resized to 32x32 before further preprocessing is applied. In the MNIST-CIF10 scenario, the MNIST dataset is converted from grayscale to RGB to ensure compatibility with the Resnet-18 model. In the CIF10-INV scenario, the colour of each pixel is inverted to produce the Inverse CIFAR-10 dataset. The final step of preprocessing re-scales the pixel values to [0,1].

Following the preprocessing, the images are then split by class into separate tasks. Each experiment run has its own random split. The exception to this is the experiments in section \ref{subsec:hierarchical_organisation}. Here, we performed the split, trained the experts, then finally applied all three methods. Thus, the generated trees can be directly compared as the splits (and experts) are identical.

\textbf{Base Models}
All models were implemented using PyTorch. In the PNIST, SMNIST and MNIST-KMNIST scenarios, each expert was a 3-layer MLP with a 4-layer MLP as the autoencoder. The MLP base model consists of 3 Linear() layers followed by a ReLU() each. The hidden activations were 100-dimensional. We treat the final outputs as logits and train them using CrossEntropyLoss(). During test time, the prediction is simply the index with the highest logit. We used a batch size of 128 in all cases.

The MLP autoencoder is a variational autoencoder with a 32-dimensional latent space. The encoder layers, in order, are Linear(784,512), ReLU(), then a separate Linear(512,32) for the mean and log-variance. The latent distribution is sampled using the reparametrisation trick. Let $\epsilon$ be the 32-dimensional vector sampled from a Gaussian distribution with 0 mean and 1 variance. The latent sample is then $\mu + e^{(\sigma*0.5)} * \epsilon$ where $\mu$ and $\sigma$ are the mean and log-variance vectors outputted by the encoder. At this point, we also compute the KL divergence loss of the output mean and log-variance as $-0.5 * (1 + \sigma - \mu^{2} - e^{\sigma})$. To obtain a scalar for the KL loss, we sum across the 32 dimensions.

The decoder layers are Linear(32,512), ReLU(), Linear(512, 784), Sigmoid(). To train the autoencoder, we combined the computed KL loss with the MSELoss() between the decoder output and the input image Tensor.

In all other scenarios, we use Resnet-18 as the base model and a convolutional variational autoencoder with 64-dimensional latent space. For reference, the arguments to Conv2d, in order, are input channels, output channels, kernel size, stride and padding. The encoder layers are Conv2d(3,32,3,1,1), BatchNorm2d(32), ReLU(), Conv2d(32, 64, 3, 1, 1), BatchNorm2d(64), ReLU(), Flatten(), Linear(65536, 128), followed by a separate Linear(128, 64) for the mean and log-variance.

The decoder layers are Linear(64, 128), ReLU(), Linear(128, 65536), ReLU(), Unflatten(64Cx32Hx32W), ConvTranspose2d(64, 32, 3, 1, 1), BatchNorm2d(32), ReLU(), ConvTranspose2d(32, 3, 3, 1, 1). As with the fully connected autoencoder, we use MSELoss() for the reconstruction error and the same calculation for the KL loss.

We tested regular autoencoders in place of variational autoencoders by simply using the encoder mean with no sampling, but found the results to be worse in all cases. We also experimented between SGD and Adam as the optimiser. In the CIF100, ImgNet and CIF-ImgNet scenarios, we used Adam with lr=0.001 and weight decay=0.0001. In all other cases, we used SGD with lr=0.01, momentum=0.9 and weight decay=0.0001

\textbf{Hyperparameters}
Each experiment was repeated 5 times. The following hyperparameter values were used in all scenarios:
\begin{itemize}
    \item High-Loss Threshold $\epsilon=4$
    \item Z-score threshold $\epsilon_{review} = 20$
    \item Promotion\_path\_threshold = 0.98
    \item High-Loss buffer capacity = 20
    \item Expert replay buffer capacity = 10
\end{itemize}

The following hyperparameters were used per scenario:
\begin{itemize}
    \item PMNIST $\epsilon_{promotion} = 0.98$, epochs = 40
    \item SMNIST $\epsilon_{promotion} = 0.98$, epochs = 40
    \item CIF10 $\epsilon_{promotion} = 0.95$, epochs = 200
    \item CIF100(10) $\epsilon_{promotion} = 0.75$, epochs = 400
    \item CIF100(20) $\epsilon_{promotion} = 0.80$, epochs = 400
    \item ImgNet(10) $\epsilon_{promotion} = 0.5$, epochs = 400
    \item ImgNet(20) $\epsilon_{promotion} = 0.65$, epochs = 400
    \item MNIST-KMNIST $\epsilon_{promotion} = 0.98$, epochs = 100
    \item MNIST-CIF10 $\epsilon_{promotion} = 0.98$, epochs = 200
    \item CIF10-INV $\epsilon_{promotion} = 0.95$, epochs = 200
    \item CIF-ImgNet $\epsilon_{promotion} = 0.5$, epochs = 400
\end{itemize}
\end{document}